%% file: main_arxiv.tex
\ifdefined\XeTeXversion\else
  \ifdefined\pdfoutput\pdfoutput=1\fi
\fi
\documentclass{article}
\usepackage{arxiv_default}
\input{preamble}

\title{DRD-MOPD: Domain-Routed Delta Optimization\\for Multi-Teacher On-Policy Distillation}
\author{Lin Yao$^{1,2}$\\
$^1$School of Computer Science, Shanghai Jiao Tong University, Shanghai, China\\
$^2$Zhongguancun Academy, Beijing, China\\
\texttt{lin.yao@sjtu.edu.cn}}
\date{}
\begin{document}
\maketitle
\input{body}
\bibliographystyle{plainnat}
\bibliography{refs}
\clearpage
\appendix
\input{appendix}
\end{document}

%% file: preamble.tex
\ifdefined\XeTeXversion\else\usepackage[utf8]{inputenc}\fi
\usepackage[T1]{fontenc}
\usepackage{amsmath,amssymb,amsthm}
\usepackage{booktabs,array,tabularx}
\usepackage{graphicx,xcolor}
\usepackage{enumitem}
\usepackage{xurl}
\usepackage{hyperref}
\hypersetup{hidelinks,pdftitle={DRD-MOPD: Domain-Routed Delta Optimization for Multi-Teacher On-Policy Distillation},pdfauthor={Lin Yao}}
\newcommand{\sg}{\operatorname{sg}}
\newcommand{\clip}{\operatorname{clip}}

\newcommand{\own}{\textnormal{\textsc{Own}}}
\newcommand{\unavailable}{---}

%% file: body.tex
\begin{abstract}
Multi-teacher on-policy distillation can yield uneven domain capabilities.
We propose Domain-Routed Delta Optimization (DRD-MOPD), composing a frozen
base with expert-initialized trainable deltas. Domains update their deltas
with independent optimizer states and clipping. After joint training,
Post-separate exports the base plus one trained delta per domain, selected
by known task domain without retraining. Using AdamW, the unweighted full student
reaches 32.83\% Macro across six benchmarks after 20 updates, versus
32.21\% for a matched shared control. Post-separate raises Macro with
Open-MOPD weighting.
\end{abstract}

\section{Introduction}
Multi-teacher on-policy distillation (MOPD) aims to consolidate domain
experts into one generalist student~\citep{ma2026mopd}. It builds on
knowledge distillation~\citep{hinton2015distilling,rusu2016policy} and uses
teacher token distributions to supervise responses generated by the
student itself~\citep{agarwal2024gkd,gu2024minillm}. Integrating these
signals does not necessarily recover each teacher's capabilities evenly.
Open-MOPD identifies substantial cross-domain imbalance, particularly
for short-response instruction-following tasks, and addresses it through
token-share balancing and gap-following allocation~\citep{gao2026openmopd}.
These mechanisms explicitly adjust the relative contributions of the
domains to the training loss.

This motivates a direct question: can the optimizer itself handle domain
balancing? Adaptive optimizers already use gradient history to adjust
parameter-update scales. Applying this mechanism independently to each
domain lets its updates adapt to its own gradient statistics. We study
domain-wise adaptive optimization as a way to simplify the
balancing design in MOPD, while preserving expert components that remain
separable and composable after training.

We introduce Domain-Routed Delta Optimization (DRD-MOPD). Starting from a
common base, task arithmetic represents each expert by its weight
difference from that base~\citep{ilharco2023task}. We retain these differences
as full-size trainable components $P_j$ and keep the base $B$ frozen,
forming a single dense student $W=B+\sum_{j=1}^{K}P_j$. Their initial sum is
the task-arithmetic merge with unit coefficients. In \own{}, prompts from
all domains are sampled in equal numbers, each response is supervised by
its domain teacher, and its gradients update the corresponding component.
Each component maintains independent adaptive optimizer state and receives
independent gradient clipping. All components participate in the same
student forward pass. The formulation applies to optimizers whose
gradient-history state and adaptive update scales can be maintained per
component; we instantiate it with AdamW in the experiments.

The retained components also provide useful outputs beyond the combined
student. After joint training, the complete sum can be deployed as one
dense model, or the base and a selected component can be exported as a
domain model. Selected subsets and coefficients define further
recompositions using the same trained tensors. Thus, the parameter
representation used for domain-wise optimization also preserves the
pieces needed for subsequent model selection and composition. We evaluate
the full student and domain-selected exports, and use geometric analysis
to characterize how the original expert directions evolve during training.

In the three-domain Open-MOPD setting, Ours without additional domain-loss
weighting achieves 32.83\% Macro after 20 updates. Shared only, a shared
student with the same initialization, sampling, weighting, and update
budget, achieves 32.21\%. The 0.62-point difference is concentrated in
Code (+1.17 points) and IF (+0.61), with a smaller Math difference (+0.08).
Relative to the TA initialization (32.44\%), however, Ours improves IF
while slightly reducing Math and Code. These are distinct comparisons:
the shared control assesses the combined optimization design, whereas TA
measures the change from the starting model.

Retaining Open-MOPD's weighting in Open-MOPD + Ours gives 32.72\% Macro
for the full student and 33.06\% for its domain-selected exports.
Unweighted Ours instead moves from 32.83\% to 32.60\% after separation.
Thus, the same component representation supports full and domain-selected
deployment, but the preferred output depends on the training configuration
and domain. Section~\ref{sec:ablation} compares all four training
configurations and their six evaluated outputs.
Sections~\ref{sec:weight-analysis} and~\ref{sec:training-dynamics} provide
complementary geometry and training diagnostics without treating them
as direct measures of capability retention.

Our contributions are threefold. First, we formulate domain-wise adaptive
optimization for MOPD, using independent component updates to investigate
balancing through the optimizer. Second, we retain expert-initialized
parameter components throughout joint training, supporting both a unified
student and separable, recomposable domain models. Third, we instantiate
the framework with AdamW, compare it with shared optimization under
matched sampling and weighting, and evaluate full and domain-selected
outputs with and without Open-MOPD's weighting. Optimization analysis and
geometric measurements characterize the update rules and component evolution.

\section{Related Work}
\label{sec:related}
\paragraph{Knowledge and on-policy distillation.}
Knowledge distillation compresses teacher or ensemble predictions into one
student~\citep{hinton2015distilling}. Policy Distillation extends this idea to
consolidating task-specific reinforcement-learning policies~\citep{rusu2016policy}.
For language models, Generalized Knowledge Distillation (GKD) trains on
student-generated sequences to address the
mismatch between training and inference prefixes~\citep{agarwal2024gkd}.
MiniLLM optimizes reverse KL through on-policy updates~\citep{gu2024minillm}.
These works establish the supervision and sampling principles underlying
on-policy capability transfer.

\paragraph{Multi-teacher on-policy distillation.}
MOPD first trains specialized domain teachers through reinforcement learning,
then distills their token distributions into one student on its own generated
trajectories, assigning each trajectory to its domain teacher~\citep{ma2026mopd}.
Open-MOPD studies uneven recovery and introduces token-share balancing,
gap-following allocation, and student reward refresh~\citep{gao2026openmopd}.
We investigate optimizer-based domain balancing in this multi-teacher
on-policy setting, while retaining the expert parameter components used
to initialize and compose the student.

\paragraph{Model merging and task arithmetic.}
Model soups average models fine-tuned from a shared starting point into one
set of weights~\citep{wortsman2022soups}. Task arithmetic defines task vectors
as weight differences from a common reference and composes them through
addition, scaling, and subtraction~\citep{ilharco2023task}.
TIES-Merging addresses interference by trimming small changes and resolving
sign conflicts before merging~\citep{yadav2023ties}. Work on tangent-space
task arithmetic relates successful composition to weight disentanglement and
localized functional effects of task directions~\citep{ortiz2023tangent}.
Our initialization uses task arithmetic with unit coefficients; subsequent
training updates the full increment tensors.

\paragraph{Parameter-space MOPD.}
IM-MOPD initializes from a merge and periodically injects task-vector increments
for domains with insufficient recovery~\citep{kim2026immopd}.
Our formulation fixes the base and updates the full expert-increment tensors
through domain-routed gradients, maintaining a separate update history
for each increment.

\paragraph{Task-specific optimization and modular distillation.}
PCGrad modifies conflicting task gradients by projection in the shared
parameter space~\citep{yu2020pcgrad}. AdaTask applies task-specific gradient
statistics to adaptive optimizers, including AdaGrad, RMSProp, and Adam,
and combines their updates to shared parameters~\citep{yang2023adatask}.
We bring domain-wise adaptive optimization into expert-initialized
multi-teacher on-policy distillation, retaining parameter components for
subsequent domain-model export and recomposition. Appendix~\ref{app:proofs}
gives the task-wise Adam equivalence for \own{} under matched clipping,
state histories, and weight-decay semantics.
Poly-OPD uses a frozen image-generation backbone, shared attention LoRA, and
capability-specific FFN adapters under multi-teacher OPD~\citep{fu2026polyopd}.
Its selectable adapters differ from our dense student, which uses the
sum of all components during training and Full inference. Finally,
$\Delta$-MOPD composes teacher-minus-base \emph{logit}
shifts into a distillation target~\citep{xu2026deltamopd}; our trainable components are
\emph{parameter} tensors, and the two approaches operate in different spaces.

\section{Method}
\label{sec:method}

\subsection{Student parameterization}
\label{sec:setup}
Consider $K\geq2$ domains indexed by $j\in\{1,\ldots,K\}$. Let $B$ denote
the shared base parameters and $E_j$ an initialization expert for domain
$j$, with parameters $W_{E_j}$. The experts share the parameter layout of $B$.
For each domain, we maintain a full-parameter component $P_j$, initialized
from the corresponding expert-minus-base weights:
\[
 P_{j,0}=W_{E_j}-B.
\]
The base $B$ remains frozen. Training updates complete component tensors,
with each domain updating its corresponding component.
The student parameters are their sum,
\begin{equation}
 W=B+\sum_{j=1}^{K}P_j.
 \label{eq:composition}
\end{equation}
At initialization, this gives the task-arithmetic merge of the $K$ experts
with unit coefficients~\citep{ilharco2023task}.

In training and default composed inference, every input is processed by
the full dense network parameterized by $W$.
All components contribute to response generation and optimization during
training and to Full inference, including components that receive no update from
that sample. The decomposition introduces no separate forward branches or
input-adaptive gating. It separates parameter-update histories within one
shared network; the default inference model requires only the composed weights $W$.
After joint training, \emph{Post-separate} exports one dense model per
domain, $W_j=B+P_j$, by adding only the corresponding trained component
to the frozen base. At inference, the known task domain $j$ selects $W_j$;
the other components are omitted. All exports reuse the same joint
training checkpoint and require no additional training. Retaining the
components also permits recompositions $W(\alpha)=B+\sum_j\alpha_jP_j$
with chosen coefficients. Section~\ref{sec:reference-results} evaluates
Full and Post-separate.

\subsection{Domain-supervised distillation}
\label{sec:shared-procedure}
Each domain $i$ is assigned a frozen supervising teacher $T_i$. This teacher
need not be the initialization expert $E_i$, and multiple domains may share
a supervising checkpoint. Let $\pi_W$ and $\pi_{T_i}$ denote the conditional
token policies of the student and teacher, respectively.

In \own{}, each rollout batch contains equal numbers of prompts from all
domains.
For each prompt $x$, the student generates one response
$y\sim\pi_W(\cdot\mid x)$ by stochastic autoregressive sampling. A rollout
step denotes the generation of this batch of responses. For a sample from
domain $i$, teacher $T_i$ provides token probabilities at each sampled
prefix $(x,y_{<\ell})$, where $\ell$ indexes response tokens. The
token-level matching target is the reverse KL
\[
 \operatorname{KL}\!\left(
 \pi_W(\cdot\mid x,y_{<\ell})\,\middle\|\,
 \pi_{T_i}(\cdot\mid x,y_{<\ell})\right).
\]
The teacher therefore supplies supervision on the student's own trajectory.
The baseline loss uses teacher--student log-probability differences to
construct distillation advantages, which are treated as constants when
differentiating a clipped policy surrogate. Domain weights determine the
relative loss contributions. The loss approximation and weighting scheme
are specified in Appendix~\ref{app:config}.

\subsection{Parameter-update routing}
\label{sec:modes}
Let $i$ index the sample's domain and $j$ index a parameter component $P_j$,
with $i,j\in\{1,\ldots,K\}$. In \own{}, a batch contains all domains,
and domain-$i$ losses are routed to $P_i$ (Table~\ref{tab:modes}).
Each component receives only its corresponding domain's gradients.

\begin{table}[!htbp]
\centering\small
\caption{Correspondence between sample domain $i$ and updated parameter
component $P_j$.}
\label{tab:modes}
\begin{tabular}{@{}lll@{}}
\toprule
Method & Sample domain & Updated component $P_j$\\
\midrule
\own & All domains $i$ & $j=i$ for each sample\\
\bottomrule
\end{tabular}
\end{table}

\own{} retains all components in every forward pass and uses the
distillation objective in Appendix~\ref{app:config}.
Figure~\ref{fig:routing} shows its update structure for $K=3$.

\begin{figure}[!htbp]
\centering
\includegraphics[width=\linewidth]{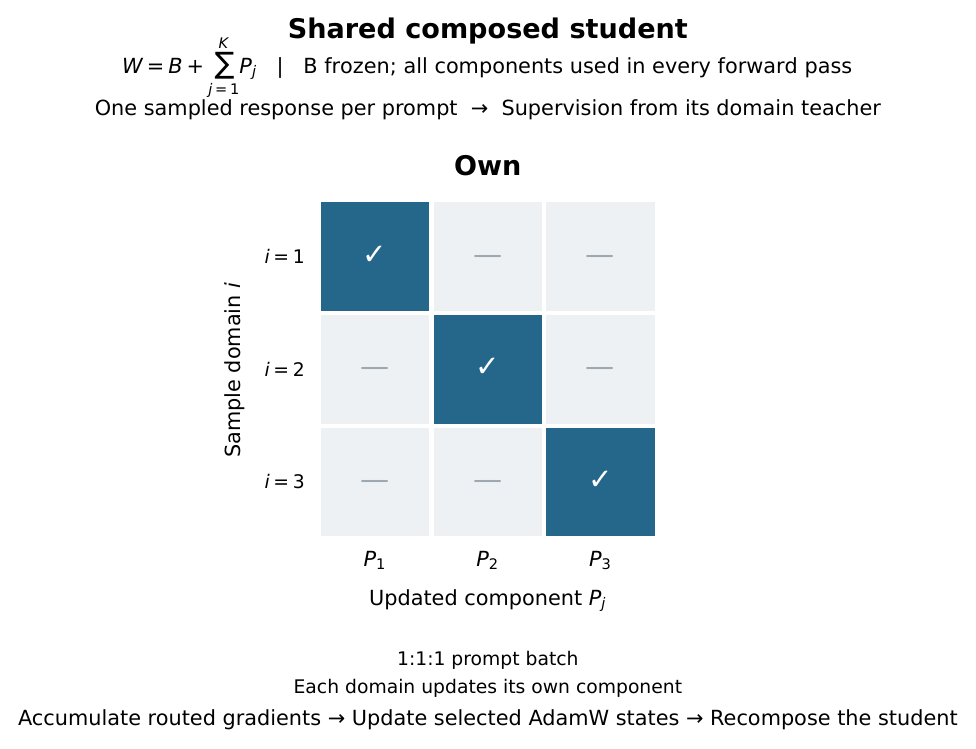}
\caption{Domain-to-component gradient routing for $K=3$.
Rows denote sample domains and teachers; columns denote parameter components.
Check marks select gradient recipients. All components participate in
every forward pass.}
\label{fig:routing}
\end{figure}

\subsection{Optimization}
\label{sec:implementation}
\label{sec:analysis}
\label{sec:geometry}
Each component maintains its own optimizer state. Let $s_{j,t}$ denote that
state and $g_{j,t}$ the gradient accumulated for an active component in
optimizer step $t$. With $C_\tau$ denoting L2 gradient clipping at threshold
$\tau$, a stateful optimizer $\mathcal{U}$ updates the component as
\[
 (P_{j,t+1},s_{j,t+1})
 =\mathcal{U}\bigl(P_{j,t},C_\tau(g_{j,t}),s_{j,t}\bigr).
\]
Adaptive optimizers can use each component's gradient history to determine
its update scales independently. This interface separates the component
representation and routing rule from the choice of adaptive optimizer.
Our implementation instantiates $\mathcal{U}$ with AdamW, maintaining
independent first and second moments and an update counter, all initialized
to zero~\citep{loshchilov2019adamw}.
Within an optimizer mini-batch, the weighted losses are differentiated at
the same current $W$, and their gradients are accumulated according to the
routing rule before any parameters change. Each selected component then
receives one AdamW update after independent L2 gradient clipping.
Unselected components retain their parameters, moments, and counters and
receive no weight decay. The updated components are added to the frozen
base according to Eq.~\eqref{eq:composition}, yielding the student used
for subsequent computation.

Optimizer states persist across rollout steps.
If a rollout batch is split into multiple optimizer mini-batches,
the sampled responses and frozen-teacher targets are reused, while the
student-dependent distillation advantages can be refreshed at the current
parameters before each update when reward refresh is enabled.
Appendix~\ref{app:distributed} gives the
gradient-accumulation and state semantics.

At a fixed $W$ and batch, plain SGD without clipping or weight decay makes
\own{} equivalent to applying the summed domain gradients directly to the
student. Independent
Adam states generally break this equivalence to a shared optimizer.
Under matching clipping and state histories with zero weight decay,
\own{} instead corresponds to task-wise Adam.
Separate component histories do not imply independent domain behavior:
every component update changes the shared $W$ used by all domains.
Appendix~\ref{app:proofs} derives these relationships.

\section{Results}
\label{sec:experiments}
\subsection{Experimental setting}
\label{sec:experimental-setting}
We use the Open-MOPD setting~\citep{gao2026openmopd} with Math, Code,
and IF ($K=3$). The base $B$ is SmolLM3-3B MixSFT; the original RL-Math,
RL-Code, and RL-IF increments initialize the three components. Their
unit-coefficient sum gives the TA $[1,1,1]$ initial student~\citep{ilharco2023task}.
Math and Code are supervised by a frozen copy of this TA mixture,
while IF uses the frozen original RL-IF expert. Thus three experts
initialize the components, and two distinct checkpoints provide supervision.
This teacher choice is fixed across our runs and is not separately ablated.

Ours uses equal domain quotas and independent AdamW states and clipping,
without Open-MOPD's additional domain-loss weighting. All local Ours and
strategy-ablation checkpoints are evaluated after 20 optimizer updates;
the released Open-MOPD Final checkpoint is at step 200.
Training settings and model sources are in Appendix~\ref{app:config}.
In Table~\ref{tab:capability}, Open-MOPD denotes the locally evaluated
released Final model. In Table~\ref{tab:ablation} and the training-dynamics
figures, it denotes our 20-update shared control using Open-MOPD's
strategies. Shared only is a separate control with equal sampling and
unit domain-loss weights.

Evaluation covers AIME 2024/2025, LiveCodeBench v5/v6, IFEval, and IFBench
test. Domain scores average their two benchmarks; Macro averages the
three domains. All main-table scores use our local benchmark, decoding,
and scoring protocol. Scores, differences, and recovery rates are computed
from unrounded measurements and then rounded for display. Each benchmark
is evaluated once; generation
topologies and sampling variation limit comparisons of small differences
(Appendix~\ref{app:metrics}). Original published scores appear separately
in Appendix~\ref{app:publishedreferences}. The geometry analysis uses the
same Ours checkpoint as the capability results.

\subsection{Capability evaluation}
\label{sec:reference-results}
Table~\ref{tab:capability} compares the local baselines and our two inference
configurations from the same joint training checkpoint. Full uses
$B+\sum_jP_j$ for every domain. Post-separate exports three models,
$B+P_{\mathrm{Math}}$, $B+P_{\mathrm{Code}}$, and $B+P_{\mathrm{IF}}$,
and evaluates each only on its corresponding domain, without additional
training. Its Macro aggregates these domain scores. RouteRL likewise
selects domain-specific models; the remaining rows each use one dense model.
Following Open-MOPD~\citep{gao2026openmopd}, recovery normalizes the macro
gain between MixSFT and the original routed RL experts:
\[
 R(M)=100\,\frac{S(M)-S(\mathrm{MixSFT})}
                    {S(\mathrm{RouteRL})-S(\mathrm{MixSFT})},
\]
where $S$ is the unrounded local Macro score. Recovery can exceed 100\%
without every domain exceeding its expert.

\paragraph{Full improves IF relative to TA initialization.}
Ours (Full) scores 32.83\% Macro, compared with 32.44\% for TA.
The gain of 0.39 percentage points reflects a domain trade-off:
IF increases by 1.81 points, while Math and Code decrease by 0.47 and
0.18 points. Recovery rises from 114.02\% to 120.84\%.
The orthogonalized merge scores 32.36\% Macro, slightly below TA.
The local evaluations of the released Open-MOPD Final model and RouteRL
score 30.66\% and 31.64\%, respectively. Ours (Full) exceeds both in
each domain, but TA already exceeds both on Macro. These comparisons
include differences in initialization and training budget;
Section~\ref{sec:ablation} provides controls starting from the same TA model.

\paragraph{Post-separate changes the domain trade-off.}
Selecting one trained component per domain yields 32.60\% Macro,
0.23 points below Full. Math improves by 1.15 points, while Code and IF
decrease by 0.19 and 1.65 points. The retained components therefore
support a different inference configuration, with no aggregate gain for
unweighted Ours in this evaluation. Section~\ref{sec:ablation} also evaluates
separation for the weighted configuration. Sample counts and length-stop statistics are reported
in Appendix~\ref{app:metrics}, Table~\ref{tab:teacher-results}.

\begin{table}[!htbp]
\centering\small
\caption{Local capability scores (\%). Domain scores average two benchmarks;
Macro averages domains. Full and Post-separate use the same trained
components. Recovery uses local MixSFT/RouteRL endpoints.
Open-MOPD is the released Final checkpoint evaluated locally.
$\dagger$: our orthogonalized merge (Appendix~\ref{app:weight-audit}).}
\label{tab:capability}
\footnotesize\setlength{\tabcolsep}{2.5pt}
\input{reference_results}
\end{table}

\subsection{Ablation of Open-MOPD strategies and domain-routed optimization}
\label{sec:ablation}
Table~\ref{tab:ablation} compares four training configurations: Shared only,
a shared student using Open-MOPD's strategies, Open-MOPD + Ours, and Ours.
Shared only disables M1/M2/M4 and uses 1:1:1 domain sampling, matching Ours.
Open-MOPD + Ours adds token-share balancing (M1) and gap-following
allocation (M2) to our component-wise optimization. All four use the
same TA initialization, supervising teachers, training data, update
budget, and student-top-16 loss. Both shared students use one AdamW state
and global clipping; Open-MOPD uses 2:2:1 sampling. The component-based
variants use independent states and clipping with 1:1:1 sampling.
These local controls differ from the released Open-MOPD Final model
in Table~\ref{tab:capability}.

M3 is disabled throughout. Advantage refresh (M4) is enabled in the two
Open-MOPD configurations, but adds no computation with one update per
rollout. The two component-based configurations each evaluate Full and
Post-separate using the same trained components; mechanism marks refer
to the training run. Common hyperparameters and detailed scores are in
Appendices~\ref{app:config} and~\ref{app:ablation}.

\begin{table}[!htbp]
\centering\small
\setlength{\tabcolsep}{4pt}
\caption{Strategy ablation and post-separate evaluation (scores in \%).
Check marks indicate enabled mechanisms. Sampling is 1:1:1 unless 2:2:1 is checked.
Ours repeats Table~\ref{tab:capability}. Shared only disables M1/M2/M4;
Open-MOPD denotes the local shared control with those mechanisms enabled.
$\dagger$: advantage refresh adds no computation with one update
per rollout.}
\label{tab:ablation}
\input{ablation_results}
\end{table}

\paragraph{Shared optimization with matched sampling and weighting.}
Shared only scores 25.42\% on Math, 22.58\% on Code, and 48.64\% on IF,
for 32.21\% Macro. Ours (Full) is higher by 0.08, 1.17, and 0.61
percentage points, respectively, and by 0.62 points on Macro. This control
uses the same implementation revision and processes 5,120 prompts per
domain, matching Ours. It compares the combined effects of domain-specific
optimizer states, independent clipping, and component-wise weight decay.
Ours is higher on five of the six benchmark scores and ties AIME24.

\paragraph{The two shared controls have different domain trade-offs.}
The local Open-MOPD shared control scores 29.71\% Macro, compared with
32.21\% for Shared only. Its Code average is 14.42\%, versus 22.58\%
for Shared only, while its IF average is higher (49.43\% versus 48.64\%).
This comparison jointly changes the prompt ratio and M1/M2 weighting;
it does not identify which mechanism causes the difference.

\paragraph{Weighting changes the full student's domain trade-off.}
Ours and Open-MOPD + Ours (Full) score 32.83\% and 32.72\% Macro.
Ours is higher by 1.43 points on Math and 0.20 on Code, but lower by
1.31 on IF. Their Macro difference rounds to 0.10 points when computed
from the unrounded scores.

\paragraph{Separation helps the weighted configuration, but not unweighted Ours.}
Post-separate raises Open-MOPD + Ours from 32.72\% to 33.06\% Macro,
the highest Macro among the six outputs in Table~\ref{tab:ablation}.
Relative to its own Full model, Math increases by 1.30 points and Code
by 0.79, while IF decreases by 1.07. For unweighted Ours, separation
instead lowers Macro from 32.83\% to 32.60\%
(Section~\ref{sec:reference-results}). Both configurations gain Math and
lose IF after separation; Code changes in opposite directions.

Shared only removes sampling and loss-weighting differences from its
comparison with Ours, but does not isolate optimizer state from clipping
and weight-decay targets. Comparisons involving Open-MOPD additionally
change sampling and M1/M2 together. Open-MOPD + Ours also uses an earlier
implementation revision and a different generation topology. The results
do not isolate individual mechanisms or establish a significant Macro gain.

\input{weight_analysis}

\subsection{Training dynamics: teacher alignment and adaptive updates}
\label{sec:training-dynamics}
\begin{figure}[!htbp]
\centering
\includegraphics[width=\linewidth]{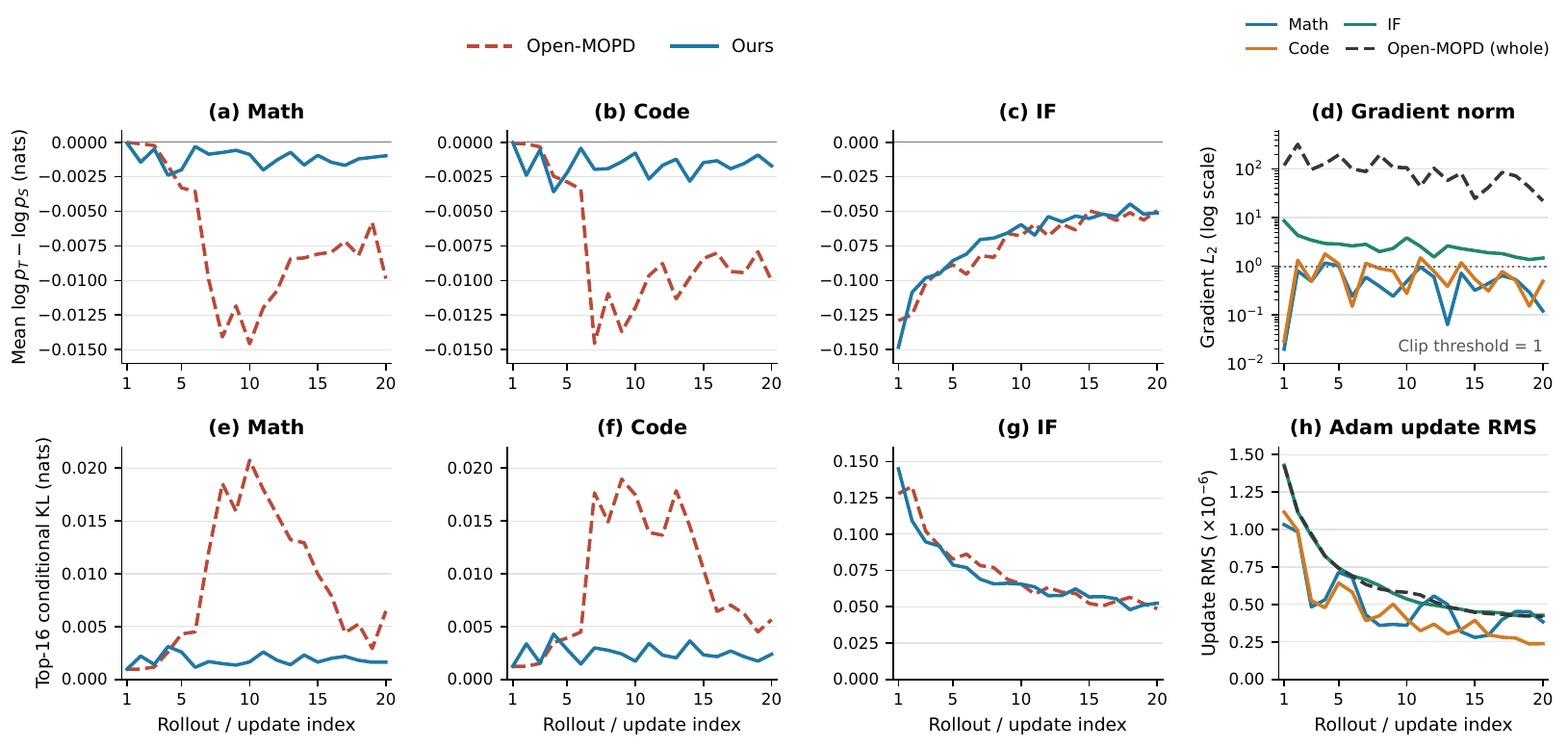}
\caption{Teacher alignment and adaptive updates for Ours and the local
shared control (Open-MOPD). (a--c) Signed sampled-token
$\log p_T-\log p_S$; (e--g) top-16 conditional reverse KL;
(d) pre-clipping gradient L2 norm; (h) gradient-driven Adam update RMS.
Panels (d,h) compare Ours domain components with shared whole-student
quantities; the dotted line in (d) marks the clipping threshold of 1.
Metric definitions are in Appendix~\ref{app:ablation}.}
\label{fig:training-dynamics}
\end{figure}

Figure~\ref{fig:training-dynamics} compares Ours with the local
Open-MOPD shared control over the first 20 updates. This is the 29.71\%
Macro configuration in Table~\ref{tab:ablation}; Shared only (32.21\%)
is not plotted. Alignment is measured before each update on the responses
sampled by that run, so the final alignment point precedes the evaluated
step-20 checkpoint. Metric definitions are given in Appendix~\ref{app:ablation}.

\paragraph{Math and Code stay closer to their supervising teacher.}
Both domains start near their TA teacher. Ours shows less drift in the
sampled-token log-probability gap and lower conditional reverse KL than
the shared control. The mean top-16 KL values for Math/Code are
0.00182/0.00247 nats for Ours, versus 0.00892/0.00925 for the shared
control. IF's log-probability gap moves toward zero in both runs.
The KL is restricted to the student's candidate set, and the sampled-token
gap is not itself a divergence; these diagnostics complement the
benchmark scores rather than measure domain capabilities directly.

\paragraph{Independent clipping permits different domain update scales.}
The shared control clips its combined gradient at every update.
Ours clips Math, Code, and IF in 2, 6, and 20 updates, respectively,
so clipping IF does not directly rescale the other components' gradients.
Despite different raw gradient norms, the mean gradient-driven Adam
update RMS values are of comparable scale:
$(0.503,0.455,0.637)\times10^{-6}$ for Math, Code, and IF.
The shared whole-student value is $0.640\times10^{-6}$ and cannot be
attributed to individual domains. Adam's history-dependent normalization
therefore matters alongside clipping.

Each trajectory comes from one run, with different prompt allocation
and sampled prefixes. The curves describe the combined training
configurations; they do not isolate optimizer state from sampling,
clipping, or loss weighting. Supplementary optimization diagnostics are
in Appendix~\ref{app:ablation}.

\clearpage
\section{Conclusion}
We introduce DRD-MOPD to study domain-wise adaptive optimization for
balancing multi-teacher on-policy distillation. The student is composed
from a frozen base and expert-initialized parameter components, each
updated from its own domain with independent optimizer state and clipping.
The framework accommodates adaptive update rules with component-specific
gradient histories, and the experiments instantiate it with AdamW.
Retaining the components supports both a unified student and
domain-selected exports from the same joint training run.

After 20 updates, unweighted Ours achieves 32.83\% Macro, compared with
32.21\% for Shared only under matched sampling and weighting and 32.44\%
for the TA initialization. The comparison with Shared only favors Ours
in all three domain averages; the comparison with TA instead shows an
IF gain with small Math and Code losses. Open-MOPD + Ours achieves
32.72\% as a full student and 33.06\% after separation, while unweighted
Ours decreases to 32.60\% after separation. The benefit of selecting a
single component therefore depends on the training configuration.

In the joint projection onto the initial expert basis, each component
retains a near-unit coefficient on its own initial direction. IF has
the largest rotation, and new changes extend outside the initial expert
span. Together with the training
diagnostics, they characterize how the components evolve without
establishing a causal explanation of the benchmark differences.
The current evidence supports a composable parameter representation and
a modest observed gain for the combined domain-wise optimization design;
isolating its mechanisms and establishing stability require further experiments.

\section{Limitations}
Shared only matches Ours' initialization, teachers, sampling, weighting,
implementation revision, and 20-update budget, but optimizer states,
clipping, and weight-decay targets change together. The observed
0.62-point Macro gain therefore concerns this combined design.
Comparisons with the local Open-MOPD shared control additionally change
sampling and M1/M2, while Open-MOPD + Ours uses an earlier implementation
revision. Separate M1/M2 effects and active multi-update M4 remain
unresolved. The experiments use AdamW; other adaptive optimizers have
not been evaluated.

Each ablation configuration has one training run; each model is evaluated
once per benchmark.
The main table uses a common local benchmark, decoding, and scoring
protocol, but generation topologies and unfixed generation randomness
limit comparisons of small differences. The four ablation runs share
the distillation objective and 20-update budget; comparisons with the
released Open-MOPD Final model additionally differ in initialization,
training protocol, and budget. The results do not establish statistical
significance, convergence, or relative training efficiency. Published
reference scores are kept separately in the Supplementary Information.

Post-separate uses three models and known domain labels. Its 33.06\%
Macro for Open-MOPD + Ours is the highest in the ablation, but the same
operation reduces unweighted Ours from 32.83\% to 32.60\%. This is not
evidence for a general benefit of separation. Each comparison uses the
same trained components, but generation samples are not paired; arbitrary
subsets and recomposition coefficients have not been evaluated.

The geometric analysis uses the same Ours checkpoint as the main
capability results and the span of three complete initial expert increments.
Near-unit projection coefficients and near-orthogonal component angles
do not establish functional independence or capability preservation.
FP32 components and BF16 exports also give different geometric
measurements. Training-dynamics figures compare Ours with the weighted
Open-MOPD shared control, not Shared only, and cannot explain the matched
control's 0.62-point difference on their own.

Full-parameter components and independent optimizer moments increase
training memory and checkpoint storage; their cost has not been
benchmarked against the shared control. The implementation covers one
3B model family and three domains with FSDP1. Math and Code share a frozen
TA supervising teacher, while IF uses the original RL-IF expert; this
teacher assignment is not ablated. Longer runs, more domains, other model
scales, and other sharding schemes remain to be evaluated.

%% file: reference_results.tex
\begin{tabular}{@{}l*{11}{r}@{}}
\toprule
 & \multicolumn{3}{c}{Math (avg@64)} & \multicolumn{3}{c}{Code (avg@10)} & \multicolumn{3}{c}{IF} & & \\
\cmidrule(lr){2-4}\cmidrule(lr){5-7}\cmidrule(lr){8-10}
Method / model & AIME24 & AIME25 & Avg & LCB v5 & LCB v6 & Avg & IFEval & IFBench & Avg & Macro & Recovery\\
\midrule
\multicolumn{12}{l}{\textit{Initial models}}\\
MixSFT (epoch 4)~\citep{gao2026openmopd} & 15.78 & 19.58 & 17.68 & 14.97 & 19.20 & 17.09 & 67.10 & 19.00 & 43.05 & 25.94 & 0.00\\
RouteRL (routed teachers)~\citep{gao2026openmopd} & 25.89 & 24.90 & 25.39 & 22.69 & 22.29 & 22.49 & 73.75 & 20.33 & 47.04 & 31.64 & 100.00\\
\midrule
\multicolumn{12}{l}{\textit{Distilled baseline}}\\
Open-MOPD~\citep{gao2026openmopd} & 21.98 & 23.39 & 22.68 & 21.20 & 22.29 & 21.74 & 72.09 & 23.00 & 47.54 & 30.66 & 82.72\\
\midrule
\multicolumn{12}{l}{\textit{Merged models constructed in this work}}\\
TA [1,1,1]~\citep{ilharco2023task} & 26.30 & 25.62 & 25.96 & 24.25 & 23.60 & 23.93 & 73.20 & 21.67 & 47.43 & 32.44 & 114.02\\
Orthogonalized merge$^\dagger$ & 23.33 & 26.46 & 24.90 & 24.85 & 23.49 & 24.17 & 73.01 & 23.00 & 48.01 & 32.36 & 112.55\\
\midrule
\multicolumn{12}{l}{\textit{Domain-routed parameter methods}}\\
Ours (Full) & 25.05 & 25.94 & 25.49 & 23.95 & 23.54 & 23.75 & 74.49 & 24.00 & 49.25 & 32.83 & 120.84\\
Ours (Post-separate) & 27.60 & 25.68 & 26.64 & 24.31 & 22.80 & 23.56 & 73.20 & 22.00 & 47.60 & 32.60 & 116.79\\
\bottomrule
\end{tabular}

%% file: ablation_results.tex
\begin{tabularx}{\linewidth}{@{}Xcccccc@{}}
\toprule
Mechanism / metric & \shortstack{Shared\\only} & Open-MOPD & \multicolumn{2}{c}{Open-MOPD + Ours} & \multicolumn{2}{c}{Ours} \\
\cmidrule(lr){4-5}\cmidrule(lr){6-7}
 & & & Full & Post-separate & Full & Post-separate\\
\midrule
\multicolumn{7}{l}{\textit{Open-MOPD mechanisms}}\\
2:2:1 prompt sampling (Math:Code:IF) & $-$ & $\checkmark$ & $-$ & $-$ & $-$ & $-$\\
Response-length / token-share balancing & $-$ & $\checkmark$ & $\checkmark$ & $\checkmark$ & $-$ & $-$\\
Teacher--student gap weighting & $-$ & $\checkmark$ & $\checkmark$ & $\checkmark$ & $-$ & $-$\\
Student advantage refresh$^{\dagger}$ & $-$ & $\checkmark$ & $\checkmark$ & $\checkmark$ & $-$ & $-$\\
\midrule
\multicolumn{7}{l}{\textit{Our mechanisms}}\\
Three independent domain optimizers & $-$ & $-$ & $\checkmark$ & $\checkmark$ & $\checkmark$ & $\checkmark$\\
\midrule
\multicolumn{7}{l}{\textit{Evaluation scores (\%)}}\\
AIME24 & 25.05 & 24.79 & 22.81 & 23.96 & 25.05 & 27.60\\
AIME25 & 25.78 & 25.73 & 25.31 & 26.77 & 25.94 & 25.68\\
LCB v5 & 22.75 & 13.59 & 24.13 & 24.97 & 23.95 & 24.31\\
LCB v6 & 22.40 & 15.26 & 22.97 & 23.71 & 23.54 & 22.80\\
IFEval & 73.94 & 74.86 & 75.79 & 75.97 & 74.49 & 73.20\\
IFBench test & 23.33 & 24.00 & 25.33 & 23.00 & 24.00 & 22.00\\
\textbf{Macro} & \textbf{32.21} & \textbf{29.71} & \textbf{32.72} & \textbf{33.06} & \textbf{32.83} & \textbf{32.60}\\
\bottomrule
\end{tabularx}

%% file: weight_analysis.tex
\subsection{Evolution of the domain components}
\label{sec:weight-analysis}
Table~\ref{tab:weight-geometry} follows the Ours components from TA
initialization to the evaluated checkpoint. Projection coefficients are
joint least-squares coordinates in the original, non-orthogonal expert
basis. Outside energy is the fraction of squared component norm outside
that basis's span. Definitions and precision controls are in
Appendix~\ref{app:weight-audit}.

\paragraph{IF changes most; pairwise geometry remains near orthogonal.}
\label{sec:own-changes}
IF rotates by $36.85^\circ$ and increases its L2 norm by 24.96\%,
compared with $22.26^\circ$/8.23\% for Math and $11.10^\circ$/2.30\% for
Code. Its outside fraction is also largest, at 35.94\%.
Each component retains a projection coefficient close to one on its own
initial direction; coefficients on other initial directions have magnitude
at most 0.0073. Despite these individual changes, the pairwise angles
remain between $87.42^\circ$ and $92.35^\circ$, close to the initial
near-orthogonal arrangement. These observations characterize
parameter geometry; they do not establish preservation of domain
capabilities or a causal benefit of domain-routed optimization.

\begin{table}[!htbp]
\centering\small
\caption{Ours component geometry. Left: rotation from TA initialization,
joint projection coefficients, L2 norms, and outside-energy percentage.
Right: pairwise angles before and after training. Projections use the
initial expert basis; definitions are in Appendix~\ref{app:weight-audit}.}
\label{tab:weight-geometry}
\label{tab:own20}
\footnotesize
\setlength{\tabcolsep}{2.5pt}
\begin{tabular}{@{}c@{\hspace{6pt}\vrule width 0.4pt\hspace{6pt}}c@{}}
\begin{tabular}[t]{@{}lrlrrr@{}}
\toprule
\multicolumn{6}{@{}l}{\textit{Ours components relative to TA initialization}}\\
\midrule
Domain & \shortstack{Angle to\\TA init. ($^\circ$)}
& \shortstack{Projection coefficients\\(Math, Code, IF)}
& \shortstack{TA init.\\$L_2$ norm} & \shortstack{Ours\\$L_2$ norm}
& \shortstack{Outside energy\\(\%)}\\
\midrule
Math & 22.26 & (1.0015, 0.0037, 0.0001) & 0.819420 & 0.886860 & 14.3486\\
Code & 11.10 & (0.0007, 1.0039, 0.0002) & 1.215318 & 1.243330 & 3.7047\\
IF & 36.85 & (-0.0035, -0.0073, 1.0000) & 0.627739 & 0.784409 & 35.9447\\
\bottomrule
\end{tabular}
&
\begin{tabular}[t]{@{}lrr@{}}
\toprule
\multicolumn{3}{@{}l}{\textit{Pairwise angles}}\\
\midrule
\shortstack{Domain\\pair} & \shortstack{TA init.\\($^\circ$)} & \shortstack{Ours\\($^\circ$)}\\
\midrule
Math--Code & 88.80 & 87.42\\
Math--IF & 89.99 & 92.35\\
Code--IF & 89.98 & 91.05\\
\bottomrule
\end{tabular}
\end{tabular}
\end{table}

%% file: appendix.tex
\section{Distributed Loss Normalization}
\label{app:distributed}
The implementation materializes the complete student weights $W$ by adding
all parameter components to the base and performs ordinary dense forward
and backward passes on this model. After capturing the gradient with respect
to $W$, the optimizer accumulates it into the selected components' gradient
buffers and rematerializes $W$ after their updates. Routing does not alter
the execution path through the network layers.

Domain routing changes micro-batch grouping while retaining Open-MOPD's
clipped policy surrogate~\citep{gao2026openmopd,schulman2017ppo}.
Let $\ell_i$ be the sum of inherited token losses for sample $i$.
For an original micro-batch indexed by $m$, let $N_m$ be its effective
response-mask sum, including the denominator's numerical stabilizer.
A sample in that micro-batch contributes
$\ell_i/N_m$.

A domain-homogeneous packed micro-batch may collect samples from several
original micro-batches. Let $m(i)$ identify sample $i$'s original micro-batch.
For a packed micro-batch indexed by $p$, let $N_p$ be the corresponding
effective response-mask sum, with the same numerical stabilizer.
Multiply each contained sample's advantages by the normalization factor
\begin{equation}
 \kappa_i=\frac{N_p}{N_{m(i)}}.
 \label{eq:normalization}
\end{equation}
The inherited surrogate is positively homogeneous in its advantages,
including the negative-advantage clipping branch. The rescaling therefore
changes the sample numerator to $\kappa_i\ell_i$, and division by $N_p$
recovers $\ell_i/N_{m(i)}$ exactly in real arithmetic. The original gradient
accumulation factor and rank averaging are retained. This corrects the
regrouping denominator while preserving teacher and question weights.
Original response masks and any per-question format masks determine both
denominators.

Distributed workers process domains in a common order, using zero-weight
padding when a worker has fewer samples. Component gradient norms are
aggregated across shards before independent clipping.

An active component is one selected by a processed domain, even when its resulting
gradient is numerically zero. Such a component follows AdamW's normal momentum and
decay update. An inactive component is not selected at all and skips the update.
Thus a zero gradient and an inactive component have different state transitions.

\section{Analysis of Student Updates}
\label{app:proofs}
This appendix derives the update properties summarized in
Section~\ref{sec:analysis}, covering gradient accumulation under SGD,
independent Adam histories, and effects on the composite student.

For the derivations in this appendix, let $K\geq2$ be the number of domains
and $g_d$ the gradient contributed by domain $d$'s sample losses with respect
to the current $W$, including domain weights and normalization.
Since $W=B+\sum_jP_j$, a sample has the same gradient with respect to $W$
and its selected $P_d$.

\paragraph{Gradient accumulation under SGD.}
Assume plain SGD with common learning rate $\eta>0$, no clipping or decay,
and normalized domain gradients evaluated at the current composite.
Write $W^+$ for the composite after one update.
Accumulating gradients into components as specified in Section~\ref{sec:modes}
and substituting their SGD changes into Eq.~\eqref{eq:composition} gives
\begin{equation}
 W^+-W=-\eta\sum_d g_d.
 \label{eq:sgd-reduction}
\end{equation}
In \own{}, each domain gradient appears once in the sum of component updates.
This identity describes
plain SGD; adaptive moments and clipping change the resulting updates.

\paragraph{Independent Adam histories.}
Assume zero weight decay. Initialize a shared student to
$B+\sum_jP_{j,0}$ and give each domain an Adam state identical to its
component's state in \own{}. Match the data, teacher targets, student loss,
normalization, clipping, learning rates, and per-domain update counters.
The two implementations then have equal domain gradients and Adam
increments. Adding those increments to the shared student gives the same
parameters as recomposing the updated components. Induction extends equality to
subsequent updates when they receive the same data and rollout random
streams. This equivalence is in real arithmetic; finite-precision
materialization and reduction order can introduce numerical differences.

If all components are active with common AdamW weight-decay coefficient $\lambda$,
the combined decay term is $-\eta\lambda\sum_jP_j=-\eta\lambda(W-B)$.
An equivalent shared-student control therefore decays toward $B$. Such a control can match \own{} using per-domain
moments without storing separate component tensors.

\paragraph{Clipping before adaptive moments.}
For a gradient vector $g$, threshold $\tau>0$, and numerical stabilizer
$\epsilon_c=10^{-6}$, define
$C_\tau(g)=g\min(1,\tau/(\|g\|_2+\epsilon_c))$.
Independent clipping supplies $C_\tau(g_d)$ to each component's moments in
\own{}, whereas clipping the mixed gradient supplies
$C_\tau(\sum_dg_d)$ to a shared optimizer. These operations generally
differ even before Adam's nonlinear normalization. The parameter-update
magnitude also depends on moment history and weight decay.

\paragraph{Gradient coordinates and component overlap.}
For this analysis, let $h_d$ concatenate domain $d$'s contributions to the
$K$ components before clipping. Under \own{},
\[
 h_d=(0,\ldots,g_d,\ldots,0),
\]
where $g_d$ occupies block $d$ and every other block is zero. Contributions
from distinct domains have zero inner product because their nonzero entries
occupy different blocks. Thus, under \own{},
\begin{equation}
 \langle h_d,h_e\rangle=0,\qquad d\ne e.
 \label{eq:routing-geometry}
\end{equation}
Summing component changes maps
the separate blocks back to the same parameter coordinates in $W$.

\paragraph{Cross-domain effects through the composite.}
Let $f_x$ be a scalar prediction statistic for a prompt $x$ from any domain,
twice continuously
differentiable near $W$. If one component changes by $\delta W$, then
\[
 f_x(W+\delta W)-f_x(W)
 =\langle\nabla_W f_x(W),\delta W\rangle+O(\|\delta W\|_2^2).
\]
The effect depends on the alignment between the update and the prediction
gradient, since every domain uses the same $W$.
For an explicit example, take a scalar composite parameter $w$ and losses
$\mathcal J_1(w)=(w-1)^2/2$ and $\mathcal J_2(w)=(w+1)^2/2$.
At $w=0$, their gradients are $-1$ and $+1$. Under \own{}, their component
vectors $(-1,0)$ and $(0,+1)$ are orthogonal. Updating only the first component
with SGD gives $w^+=\eta$, increasing the second loss by
$\eta+\eta^2/2>0$. Updating both with equal weights instead leaves the
composite unchanged while the components move by $+\eta$ and $-\eta$.
The example exhibits both cross-domain loss increases and cancellation
after composition, despite disjoint gradient coordinates.

\section{Training Configuration}
\label{app:config}
Table~\ref{tab:configuration} gives the settings used for the evaluated
Ours and strategy-ablation checkpoints. All start from the TA mixture of
Open-MOPD's three SmolLM3-3B expert increments. Math and Code use the
frozen TA mixture as their supervising teacher; IF uses the frozen RL-IF
model, as specified in Section~\ref{sec:experimental-setting}.

\paragraph{Distillation objective.}
The loss uses Open-MOPD's student-top-16 approximation and clipped policy
surrogate~\citep{gao2026openmopd,schulman2017ppo}. Let $n$ index samples,
$d_n$ their domains, and $h_{n\ell}=(x_n,y_{n,<\ell})$ their sampled prefixes.
Let $S_{n\ell}$ contain the pre-update student's 16 highest-probability
tokens at that prefix. Candidate identities and teacher scores remain fixed
for the rollout batch. For $v\in S_{n\ell}$, define
\[
\begin{aligned}
 s_{n\ell v}&=\log\pi_W(v\mid h_{n\ell}),&
 t_{n\ell v}&=\log\pi_{T_{d_n}}(v\mid h_{n\ell}),\\
 \bar p_{n\ell v}&=\frac{\exp s_{n\ell v}}
 {\sum_{u\in S_{n\ell}}\exp s_{n\ell u}},&
 A_{n\ell v}&=\sg\!\left[\bar p_{n\ell v}
 (t_{n\ell v}-s_{n\ell v})\right].
\end{aligned}
\]
Here $\sg$ stops gradients. When advantage refresh is enabled,
student-dependent terms are refreshed before each inner update if a
rollout has multiple optimizer mini-batches;
with one update, the precomputed advantages already use the same student.
The advantage retains a separate value for each candidate. Only the
student weights $\bar p$ are normalized over the candidate set; $s$ and
$t$ are log probabilities under the full vocabulary distributions.

With detached reference log probability $s^{\mathrm{ref}}_{n\ell v}$,
the candidate ratio and loss are
\[
\begin{aligned}
 r_{n\ell v}&=\exp\!\left(\clip(s_{n\ell v}-s^{\mathrm{ref}}_{n\ell v},-20,20)\right),\\
 C(r,A)&=\max\{-rA,-\clip(r,0.8,1.28)A\},\\
 \phi(r,A)&=\begin{cases}
 \min\{C(r,A),-3A\},&A<0,\\
 C(r,A),&A\geq0.
 \end{cases}
\end{aligned}
\]
For the evaluated configuration (one mini-batch and one epoch per rollout),
$s^{\mathrm{ref}}=\sg(s)$, so $r=1$ numerically while its numerator remains
differentiable. Thus ratio clipping does not activate in this configuration.
With multiple mini-batches, the reference instead uses saved pre-update
student log probabilities on the same candidate support.

Let $M_{n\ell}$ be the effective response mask, including any format mask,
and $w_{d_n}$ the domain weight. Before domain packing, the loss for original
micro-batch $m$ is
\[
 \mathcal L_m=\frac{1}{N_m}\sum_{n\in m}\sum_\ell M_{n\ell}
 \sum_{v\in S_{n\ell}}\phi(r_{n\ell v},w_{d_n} A_{n\ell v}),
 \qquad N_m=\sum_{n\in m,\ell}M_{n\ell}+\epsilon_m,
\]
where $\epsilon_m$ is the mask-mean numerical stabilizer. Candidate losses
are summed, then averaged over effective response tokens within each
original micro-batch. Appendix~\ref{app:distributed} gives the packing
correction and gradient accumulation semantics. No entropy bonus or
additional reference-policy KL penalty is used.

\paragraph{Domain weights.}
Ours and Shared only set all additional domain weights to one
and disable advantage refresh. Open-MOPD + Ours retains token-share
balancing and gap-following allocation (Section~\ref{sec:ablation});
the following weighting equations describe that hybrid configuration.
Let $\mathcal D=\{1,\ldots,K\}$ be the domains present in a global rollout batch.
For $d\in\mathcal D$, let $f_d$ be the observed response-token
fraction and $b_d=(1/|\mathcal D|)/f_d$. Let $G_d$ be the mean of
$|\sum_{v\in S_{n\ell}}A^{(0)}_{n\ell v}|$ over that domain's response
tokens, where $A^{(0)}$ is the unweighted pre-update advantage.
All domain sums below range over $\mathcal D$. For the configured weighting,
\[
 \bar G=\frac{\sum_d b_dG_d}{\sum_d b_d},\qquad
 \widetilde w_d=b_d\clip\!\left(
 \frac{\max(G_d,10^{-12})}{\max(\bar G,10^{-12})},0.05,20\right),\qquad
 w_d=\frac{\widetilde w_d}{\sum_e f_e\widetilde w_e}.
\]
This uses exponent $\alpha=1$, multiplies by the gap ratio without an
initial-step anchor, and normalizes the token-weighted mean weight to one.
Weights are computed once per rollout and held fixed during inner updates.

\paragraph{Sampling and updates.}
Training samples one response per prompt with temperature 1.0, top-$p$
1.0, and no top-$k$ cutoff. Prompts are shuffled within each domain pool
and drawn cyclically. Ours, Open-MOPD + Ours, and Shared only use equal
domain quotas; the Open-MOPD shared control uses Math:Code:IF weights
of 2:2:1. Domain quotas apply
to the global batch. The rollout and optimizer mini-batch sizes are both
768, giving one update per rollout. Consequently, enabling advantage
refresh in the Open-MOPD configurations adds no extra refresh step.
The student's top-16 restriction applies to the loss, not response sampling.

\begin{table}[htbp]
\centering\small
\caption{Training settings for the evaluated checkpoints. Method-specific mechanisms are compared in Table~\ref{tab:ablation}.}
\label{tab:configuration}
\begin{tabularx}{\linewidth}{@{}lX@{}}
\toprule
Setting & Value\\
\midrule
Domains / component order & Math, Code, IF\\
Base / initialization & Frozen MixSFT; original expert-minus-base components, TA $[1,1,1]$ start\\
Evaluated checkpoint & 20 optimizer updates (one per rollout)\\
GPUs & 32 (4 nodes, 8 GPUs per node)\\
Global rollout / mini-batch & 768 / 768 prompts\\
Responses per prompt & 1\\
Question sampling & Ours, hybrid, Shared only: 1:1:1; Open-MOPD: 2:2:1\\
Actor / student / teacher micro-batch & 1 / 1 / 1 per GPU\\
Prompt / response maximum & 2,048 / 16,384 tokens\\
Generation & Temperature 1.0; nucleus probability threshold 1.0; no top-k cutoff\\
Distillation loss support & Student-only top-16 candidates\\
Policy ratio clip / negative-advantage cap & $[0.8,1.28]$ / $3$\\
Policy epochs per rollout & 1\\
Learning rate / scheduler & $1.5\times10^{-6}$ / constant\\
Adam $\beta_1,\beta_2,\epsilon_{\mathrm{Adam}}$ & $0.9$, $0.999$, $10^{-8}$\\
Weight decay / clip norm & 0.01 / 1.0\\
Token-share balancing & Disabled for Ours; equal targets for Open-MOPD + Ours\\
Gap-following allocation & Disabled for Ours; mean absolute reward, $\alpha=1$, for Open-MOPD + Ours\\
Student reward refresh & Disabled for Ours; enabled for Open-MOPD + Ours\\
Student training & FSDP, SDPA, gradient checkpointing\\
Components and Adam states & FP32\\
Seed & 42\\
\bottomrule
\end{tabularx}
\end{table}

\paragraph{Models and data.}
Initialization uses the public Open-MOPD SmolLM3-3B MixSFT and RL-Math,
RL-Code, and RL-IF checkpoints~\citep{gao2026openmopd}.
We use its released mixed-domain training data. Exact complete-chat
deduplication removes 721 examples; removing groups with conflicting
domain labels, sources, answers, or IF verification rules removes another
5,294. A split with seed 42 holds out 512 prompts per domain, leaving
79,380 training and 1,536 validation prompts. No exact prompt overlaps
with the six evaluation sets were found; this check does not cover
paraphrases. Evaluation uses the released AIME 2024/2025, LiveCodeBench
v5/v6, IFEval, and IFBench test splits from the same data release.

\section{Evaluation Protocol}
\label{app:metrics}
All local evaluations use the released benchmark inputs and scoring
routines. AIME 2024 and 2025 have 30 questions each, with 64 sampled
responses per question. LiveCodeBench v5 and v6~\citep{jain2024livecodebench}
have 167 and 175 questions, with ten responses each. IFEval~\citep{zhou2023ifeval}
and IFBench test~\citep{pyatkin2025ifbench} have 541 and 300 prompts,
respectively, with one response each and strict prompt-level scoring.

Decoding uses temperature 0.6 for Math and 1.0 for Code and IF, top-$p$
0.95, no top-$k$ cutoff, a 32,768-token context limit, and a 30,000-token
output limit. Repeated samples are separate requests, submitted in batches
of 64. Checkpoints are evaluated in BF16 without quantization. The evaluated
MixSFT export uses RoPE base $5\times10^6$; the public Final and RL experts
use $2\times10^6$. Scoring seeds for Python, NumPy, and language detection
are zero. Generated programs that fail execution retain failure grades.

TA generation uses 16 GPUs and the orthogonalized merge uses 32. Main-table
Ours and the public baselines use eight GPUs per domain; the hybrid uses
16 for Full and eight for Post-separate. No per-request generation seed
is fixed. Each score comes from one evaluation, so small differences
include sampling and batching variation. Each Full/Post-separate comparison
uses the same trained components: Full is $B+\sum_jP_j$ and the model for
domain $j$ is $B+P_j$.

Table~\ref{tab:teacher-results} gives the response counts and length-stop
statistics for the main comparisons. Length-stopped responses remain in
the score denominators.

\begin{table}[!htbp]
\centering\small
\setlength{\tabcolsep}{4pt}
\caption{Benchmark scores and length-stop counts for TA, Ours (Full),
and Ours (Post-separate, abbreviated Split).
Each configuration has 1,243 prompts and 8,101 responses;
length-stopped responses remain in the score denominators.}
\label{tab:teacher-results}
\input{teacher_results}
\end{table}

For benchmark $b$ with $N_b$ prompts and $n_b$ sampled responses per prompt,
let $i=1,\ldots,N_b$ index prompts and $q=1,\ldots,n_b$ index sampled responses.
The indicator $z_{biq}(W)$ equals one if response $q$ to prompt $i$, generated
by model $W$, satisfies benchmark $b$'s correctness criterion, and zero otherwise.
The estimates $\widehat s_b$, $\widehat s_d$, and $\widehat s_{\mathrm{macro}}$
denote benchmark, domain, and macro-average scores, respectively:
\begin{align}
 \widehat s_b(W)&=\frac{100}{N_b}\sum_{i=1}^{N_b}
                    \frac{1}{n_b}\sum_{q=1}^{n_b}z_{biq}(W),\nonumber\\
 \widehat s_d(W)&=\frac{1}{2}\sum_{b\in\mathcal B_d}\widehat s_b(W),&
 \widehat s_{\mathrm{macro}}(W)&=\frac{1}{3}\sum_d\widehat s_d(W).
 \label{eq:evaluation}
\end{align}
Here $\mathcal B_d$ contains the two benchmarks for domain $d$; all scores are
percentages. Domain scores, Macro, differences, and recovery rates are
computed before rounding; subtracting displayed scores can therefore
differ by 0.01 points from a reported difference.
We compare models within each domain; the macro score serves
as an equal-weight summary. For domain-selected exports, the corresponding
summary is $\frac{1}{3}\sum_d\widehat s_d(W^{(d)})$, using the unrounded
domain scores from the three models.
IFEval and IFBench use strict prompt-level correctness.

Recovery uses the unrounded macro scores and the definition in
Section~\ref{sec:reference-results}. The local MixSFT and RouteRL endpoints
are 25.938768\% and 31.641203\%, respectively. Recovery is an affine
transformation of Macro and is not clipped to $[0,100]$.

\section{Weight-Geometry Definitions and Controls}
\label{app:weight-audit}
We analyze the same Ours checkpoint as in Table~\ref{tab:capability}.
Composite-model increments are measured from saved BF16 weights; domain
components are measured in FP32. Differences, inner products, and norms
use FP64 arithmetic over all parameters.

\paragraph{Projection and angles.}
All measurements in Tables~\ref{tab:weight-geometry}
and~\ref{tab:composite-geometry} use the same fixed reference basis.
Let $B$ be MixSFT and $W_{E_d}$ the original RL-Math, RL-Code, or
RL-IF expert. Set $P_{d,0}=W_{E_d}-B$ and collect the flattened increments:
\[
 D=[\operatorname{vec}(P_{\mathrm{Math},0}),
    \operatorname{vec}(P_{\mathrm{Code},0}),
    \operatorname{vec}(P_{\mathrm{IF},0})].
\]
For a checkpoint $W$, its increment, projection coefficients, and residual are
\begin{equation}
 u=\operatorname{vec}(W-B),\qquad
 a=(D^\top D)^\dagger D^\top u,\qquad
 r=u-Da,\qquad \rho_\perp=\frac{\|r\|_2^2}{\|u\|_2^2}.
 \label{eq:weight-projection}
\end{equation}
The dagger denotes the Moore--Penrose pseudoinverse. The columns of $D$
are linearly independent but not exactly orthogonal. Thus $Da$ is the
orthogonal projection onto their joint span, and $a$ gives its coordinates
in the original non-orthogonal basis. These least-squares coefficients
are generally different from separate projections
$\langle D_j,u\rangle/\|D_j\|_2^2$ and cannot be interpreted as independent
per-domain energy contributions. For component rows, replace $u$ with
the flattened component. The outside fraction is undefined for $u=0$.

For nonzero vectors $v$ and $q$, angles are
$\arccos(\langle v,q\rangle/(\|v\|_2\|q\|_2))$, reported in degrees.
In Table~\ref{tab:weight-geometry}, component angles compare each trained
component with its initialization; the table also reports L2 norms and
pairwise angles between domains. Composite angles in
Table~\ref{tab:composite-geometry} use $W_{\mathrm{TA}}-B$ as the reference.

\paragraph{Supplementary composite-model geometry.}
Table~\ref{tab:composite-geometry} describes saved composite-model
increments relative to the original expert span. For Open-MOPD Final,
we project the single increment $\operatorname{vec}(W_{\mathrm{Final}}-B)$
onto the same $D$. Its three coefficients do not recover three trained
domain components inside Final, whose pairwise angles are therefore
unavailable. Ours and the released Final model have different
initializations, training protocols, and budgets
(Section~\ref{sec:experimental-setting}). These measurements do not
isolate the effect of domain-routed optimization or rank domain capabilities.

\begin{table}[!htbp]
\centering\small
\setlength{\tabcolsep}{4pt}
\caption{Supplementary geometry of BF16 increments $u=W-B$.
Angles use the TA increment as reference; projection coefficients use the
common initial expert basis. Outside energy is reported in percent.
$\dagger$: our orthogonalized merge. Training settings differ across models.}
\label{tab:composite-geometry}
\begin{tabular}{@{}llrlrr@{}}
\toprule
Vector / checkpoint & Angle reference & \shortstack{Angle\\($^\circ$)}
& \shortstack{Projection coefficients\\(Math, Code, IF)}
& $L_2$ norm & \shortstack{Outside energy\\(\%)}\\
\midrule
\multicolumn{6}{@{}l}{\textit{Composite models relative to the original TA mixture}}\\
MixSFT~\citep{gao2026openmopd} & $u_{\mathrm{TA}}$ & --- & (0.0000, 0.0000, 0.0000) & 0.000000 & ---\\
TA $[1,1,1]$~\citep{ilharco2023task} & $u_{\mathrm{TA}}$ & 0.00 & (0.9990, 0.9993, 1.0000) & 1.607181 & 0.0576\\
Orthogonal merge$^\dagger$ & $u_{\mathrm{TA}}$ & 1.66 & (0.9969, 0.9999, 0.9995) & 1.606737 & 0.0576\\
Open-MOPD Final~\citep{gao2026openmopd} & $u_{\mathrm{TA}}$ & 84.40 & (0.0258, 0.0319, 0.3489) & 1.295021 & 97.0211\\
Ours & $u_{\mathrm{TA}}$ & 14.76 & (0.9993, 1.0005, 1.0003) & 1.662509 & 6.4396\\
\bottomrule
\end{tabular}
\end{table}

\paragraph{TA construction and export precision.}
The TA construction weights are exactly $(1,1,1)$:
\[
 W_{\mathrm{TA}}^*=B+P_{\mathrm{Math},0}+P_{\mathrm{Code},0}+P_{\mathrm{IF},0}.
\]
In exact arithmetic, $\operatorname{vec}(W_{\mathrm{TA}}^*-B)=D\mathbf1$,
so its projection coefficients are exactly $\mathbf1$, even though $D$
is non-orthogonal. The actual merge accumulates in FP32 in Math, Code, IF
order and is saved in BF16. Table~\ref{tab:composite-geometry} projects the
saved increment, yielding $(0.9990,0.9993,1.0000)$ and 0.0576\% outside
energy. This finite-precision difference does not change the construction
weights; the displayed coefficients are measured after export.

\paragraph{Representation and counting controls.}
For Ours, the outside-energy fractions are 6.44\% for the BF16 export
and 12.51\% for the FP32 components summed in FP64. These are different
numerical representations of the increment. The FP32-component sum has
projection coefficients $(0.9987,1.0004,1.0003)$ and an angle of
$20.71^\circ$ to its initial sum; the BF16-export measurements in
Table~\ref{tab:composite-geometry} describe the model used for capability
evaluation.

Tables~\ref{tab:weight-geometry} and~\ref{tab:composite-geometry}
include both stored copies of the tied
embedding/output matrix. Counting this matrix once changes Final's
outside fraction from 97.02\% to 97.01\%; using only transformer layers
gives 96.99\%. For the Ours FP32-component sum, these three conventions
give 12.51\%, 12.32\%, and 12.14\%, respectively. The maximum absolute
cosine between current domain components remains below 0.046 under all
three conventions.

\paragraph{Additional training changes.}
The total increment from the base and the change since initialization
use different reference points. For a component, define
$\Delta P_j=P_j-P_{j,0}$; for the exported student, define
$\Delta W=W-W_{\mathrm{TA}}$. Applying Eq.~\eqref{eq:weight-projection}
to these changes and normalizing by their own squared norms gives
outside fractions of 99.9801\%, 99.9598\%, and 99.9602\% for Math, Code,
and IF. The summed FP32 change has 99.9996\% outside energy, and the BF16
student change has 99.9986\%. These fractions normalize the additional
update, whereas Tables~\ref{tab:weight-geometry}
and~\ref{tab:composite-geometry} normalize total increments from the base.

\paragraph{Orthogonalized merge control.}
Let $N_D$ be the diagonal matrix of column norms of $D$, and let
$C=N_D^{-1}D^\top DN_D^{-1}$. The columns of $D$ are independent, so $C$
is positive definite. Symmetric orthogonalization gives
\[
 D_{\mathrm{orth}}=DN_D^{-1}C^{-1/2}N_D,\qquad
 D_{\mathrm{orth}}^\top D_{\mathrm{orth}}=N_D^2.
\]
The control preserves each initial increment's norm while making the
three directions orthogonal before export. It uses
$W_{\mathrm{orth}}=\operatorname{BF16}
(B+\operatorname{unvec}(D_{\mathrm{orth}}\mathbf1))$,
where $\mathbf1=(1,1,1)^\top$ and $\operatorname{unvec}$ reverses flattening.
Before rounding, the effective coefficients on the original experts are
$(0.984510,0.992992,0.999525)$. Table~\ref{tab:composite-geometry} measures the
saved BF16 model, whose coefficients also reflect export rounding.

\section{Published Reference Results}
\label{app:publishedreferences}
Table~\ref{tab:publishedreferences} reproduces Open-MOPD's published
results~\citep{gao2026openmopd}. They are provided for context and are not
mixed with our local evaluations in Table~\ref{tab:capability}.
Checkpoint, runtime, generation, and scoring differences can affect the
comparison. Recovery is copied from the original table using its own
MixSFT and RouteRL endpoints; it is not recomputed from rounded scores.

\begin{table}[!htbp]
\centering\small
\caption{Original Open-MOPD scores and recovery rates (\%), reproduced
from \citet{gao2026openmopd} for reference. These use the original paper's
evaluation and recovery endpoints.}
\label{tab:publishedreferences}
\footnotesize\setlength{\tabcolsep}{2.5pt}
\input{published_reference_results}
\end{table}

\section{Strategy Ablation: Detailed Results}
\label{app:ablation}
Table~\ref{tab:ablation-benchmarks} expands the configurations in
Table~\ref{tab:ablation}. Both shared controls evaluate the complete student. Open-MOPD + Ours and Ours each compare
Full, $B+\sum_j P_j$, with Post-separate, $B+P_j$ on domain $j$.
AIME uses avg@64, LiveCodeBench uses
ten samples to estimate pass@1, and IF uses strict prompt accuracy.

\begin{table}[!htbp]
\centering\small
\setlength{\tabcolsep}{4pt}
\caption{Benchmark scores (\%) for the strategy ablation and post-separate
models. Ours reproduces the two main results in Table~\ref{tab:capability}.}
\label{tab:ablation-benchmarks}
\input{ablation_benchmarks}
\end{table}

\paragraph{Training controls.}
All four training configurations share the initialization, supervising teachers,
and hyperparameters in Table~\ref{tab:configuration}. Ours, the hybrid,
and Shared only process 5,120 prompts per domain; Open-MOPD processes 6,140 Math,
6,140 Code, and 3,080 IF prompts under its 2:2:1 sampling weights.
Shared clipping acts on the combined gradient, and shared weight decay
acts on the complete student. Component-based configurations clip each
component independently and apply decay to the components. These differences
mean that the comparison does not isolate optimizer state alone.
Shared only uses the same code revision as Ours, disables M1/M2/M4,
and evaluates all 8,101 generated responses using eight GPUs. Its six
benchmark scores are complete; no post-separate evaluation is reported
for this shared model.

\begin{figure}[!htbp]
\centering
\includegraphics[width=\linewidth]{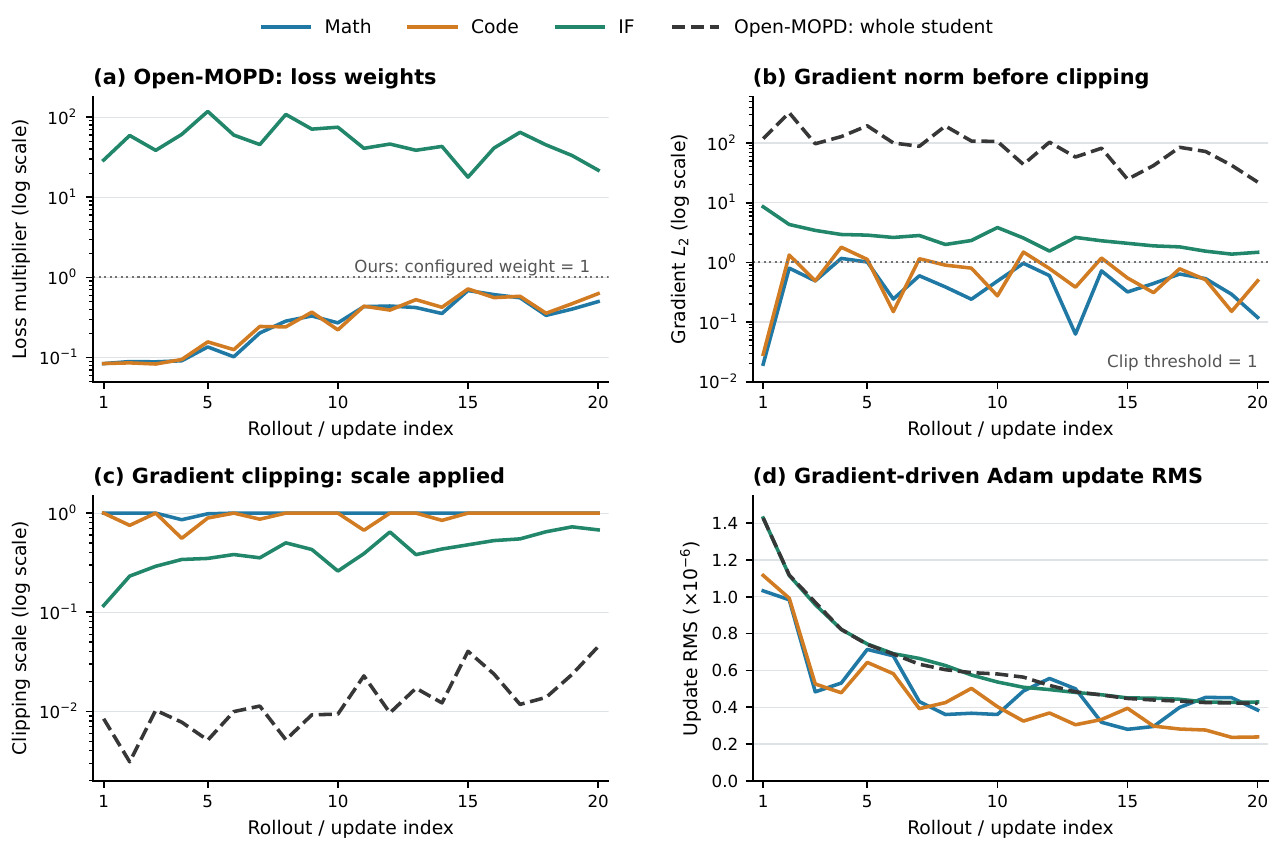}
\caption{Supplementary optimization diagnostics for Ours and the local
Open-MOPD shared control. (a) Domain loss multipliers;
(b) pre-clipping gradient norms; (c) applied clipping scales;
(d) gradient-driven Adam update RMS. In (b--d), colored lines denote Ours
components and dashed black lines the shared whole student.
The dotted unit line in (a) denotes Ours; (a--c) use log scales.}
\label{fig:training-optimization}
\end{figure}

\paragraph{Diagnostic definitions.}
Figures~\ref{fig:training-dynamics} and~\ref{fig:training-optimization}
compare Ours with the Open-MOPD shared control (29.71\% Macro), rather than
Shared only (32.21\%). They cover the first 20 updates. Step averages
weight the 20 rollouts equally. The sampled-token teacher-minus-student
log-probability gap first averages tokens within each nonempty response,
then gives responses equal weight. The conditional reverse KL renormalizes
both distributions on the student's top-16 candidate set and excludes
M1/M2 loss weights. Teacher and student use the same sampled prefixes
within each run at temperature 1.0; prefixes differ across runs. Both
alignment measures are recorded before the update, so point $t$ describes
the policy after $t-1$ updates. They do not measure full-vocabulary
divergence. All plotted points are unsmoothed.

Gradient norms are measured before clipping. The applied clipping scale
is $\min(1,1/(\lVert g\rVert_2+10^{-6}))$. Update RMS divides the L2 norm
of the gradient-driven Adam step, reconstructed from updated moments,
by the square root of the parameter count, excluding weight decay and
parameter-write rounding.
The shared diagnostics describe the whole student; Ours diagnostics
describe individual components. A per-domain shared update cannot be
recovered from the combined gradient. The hybrid is omitted from these
plots because the required diagnostics were not recorded.

\paragraph{Loss weighting, clipping, and adaptive update scales.}
Figure~\ref{fig:training-optimization} gives the full optimization diagnostics.
In the shared control, IF accounts for only 1.46\% of response tokens on
average across steps, and its M1/M2 loss multiplier ranges from 17.8 to
117.9 (mean 52.8). The combined gradient is clipped in all 20 updates,
with a mean multiplicative scale of 0.0150. In Ours, all extra domain
multipliers are one; Math, Code, and IF have mean clipping scales of
0.992, 0.929, and 0.436. A smaller clipping scale does not imply a proportional reduction in
Adam's update, whose normalization depends on its moment history.
The mean shares of the sum of component squared update norms are
30.3\%, 23.0\%, and 46.6\% for Math, Code, and IF. These shares do not
decompose the norm of the composed update, which also contains cross terms.

%% file: teacher_results.tex
\begin{tabular}{@{}lrrrrrrrr@{}}
\toprule
Benchmark & Prompts & Responses & TA (\%) & Full (\%) & Split (\%) & TA stops & Full stops & Split stops\\
\midrule
AIME24 & 30 & 1,920 & 26.30 & 25.05 & 27.60 & 120 & 202 & 199\\
AIME25 & 30 & 1,920 & 25.62 & 25.94 & 25.68 & 138 & 137 & 200\\
LCB v5 & 167 & 1,670 & 24.25 & 23.95 & 24.31 & 245 & 241 & 55\\
LCB v6 & 175 & 1,750 & 23.60 & 23.54 & 22.80 & 280 & 264 & 63\\
IFEval & 541 & 541 & 73.20 & 74.49 & 73.20 & 20 & 4 & 6\\
IFBench test & 300 & 300 & 21.67 & 24.00 & 22.00 & 18 & 7 & 8\\
\bottomrule
\end{tabular}

%% file: published_reference_results.tex
\begin{tabular}{@{}l*{11}{r}@{}}
\toprule
 & \multicolumn{3}{c}{Math (avg@64)} & \multicolumn{3}{c}{Code (avg@10)} & \multicolumn{3}{c}{IF} & & \\
\cmidrule(lr){2-4}\cmidrule(lr){5-7}\cmidrule(lr){8-10}
Method / model & AIME24 & AIME25 & Avg & LCB v5 & LCB v6 & Avg & IFEval & IFBench & Avg & Macro & Recovery\\
\midrule
MixSFT (epoch 4)~\citep{gao2026openmopd} & 15.63 & 20.26 & 17.95 & 15.99 & 19.20 & 17.60 & 66.91 & 16.00 & 41.46 & 25.67 & \unavailable\\
Naive M-OPD~\citep{ma2026mopd,gao2026openmopd} & 20.92 & 21.60 & 21.26 & 17.53 & 20.99 & 19.26 & 68.61 & 18.67 & 43.64 & 28.05 & 35.6\\
Open-MOPD~\citep{gao2026openmopd} & 21.98 & 22.86 & 22.42 & 20.84 & 22.63 & 21.73 & 74.49 & 24.67 & 49.58 & 31.24 & 83.4\\
ParamMerge-TA~\citep{ilharco2023task,gao2026openmopd} & 21.93 & 22.76 & 22.34 & 21.74 & 23.14 & 22.44 & 71.53 & 21.67 & 46.60 & 30.46 & 71.7\\
RouteOPD (routed models)~\citep{gao2026openmopd} & 22.34 & 23.96 & 23.15 & 22.28 & 21.14 & 21.71 & 75.60 & 24.00 & 49.80 & 31.55 & 88.0\\
RouteRL (routed teachers)~\citep{gao2026openmopd} & 23.65 & 24.84 & 24.24 & 22.16 & 21.31 & 21.73 & 74.49 & 27.67 & 51.08 & 32.35 & 100.0\\
\bottomrule
\end{tabular}

%% file: ablation_benchmarks.tex
\begin{tabular}{@{}l*{6}{r}@{}}
\toprule
Variant & AIME24 & AIME25 & LCB v5 & LCB v6 & IFEval & IFBench\\
\midrule
Shared only & 25.05 & 25.78 & 22.75 & 22.40 & 73.94 & 23.33\\
Open-MOPD & 24.79 & 25.73 & 13.59 & 15.26 & 74.86 & 24.00\\
Open-MOPD + Ours (Full) & 22.81 & 25.31 & 24.13 & 22.97 & 75.79 & 25.33\\
Open-MOPD + Ours (Post-separate) & 23.96 & 26.77 & 24.97 & 23.71 & 75.97 & 23.00\\
Ours (Full) & 25.05 & 25.94 & 23.95 & 23.54 & 74.49 & 24.00\\
Ours (Post-separate) & 27.60 & 25.68 & 24.31 & 22.80 & 73.20 & 22.00\\
\bottomrule
\end{tabular}